\documentclass[11pt]{article}

\usepackage[a4paper, margin=0.8in]{geometry}
\usepackage{amsmath}
\usepackage{amssymb}
\usepackage{hyperref}
\usepackage{graphicx}
\usepackage{xcolor}
\usepackage{enumitem}

\let\OLDthebibliography\thebibliography
\renewcommand\thebibliography[1]{
  \OLDthebibliography{#1}
  \setlength{\parskip}{0pt}
}

\title{The Tao of Agency\\ \Large{Autotelic AI, Embedded Agency and Dissolution of the Self}}
\author{Aritra Sarkar}
\date{}

\begin{document}

\maketitle

\begin{abstract}
Most artificial intelligence systems are built on the assumption that goals are exogenous and specified by the designer. 
Exploring what happens when an agent begins generating its own goals opens the field of autotelic AI.
Agents are expected not merely to pursue objectives but to discover them. 
In this article, we trace its consequences through intrinsic motivation, resource-driven priors, causal-interventional learning, homeostasis, and embeddedness; the last of which is found to be a necessary but not sufficient condition for autotelic agency.
Embeddedness individuates the agent at the cost of revealing that the individuation is non-unique, such that the same dynamics admit many valid partitions, each defining a different candidate self.

The deepest problem with autotelic AI is therefore not how the agent generates goals, but how it generates and relativizes the self to which the goals are assigned. 
The agent must believe in its own boundary in order to act, and see through that boundary in order to understand. 
We consolidate these developments into a single framework and extend it along three directions: a quantum formulation in which the agent-environment cut becomes physical, a philosophical reading against non-dual contemplative traditions, and a concrete LLM-based agentic instantiation.
\end{abstract}

\section{Autotelic AI}
\label{sec:goal-problem}

Artificial intelligence has been organized around a clean division of labor where the human designer specifies \textit{what} is to be achieved, while the algorithm searches for \textit{how} to achieve it. 
This separation, often traced to the agent-environment formulation \cite{russell1995modern}, has proven extraordinarily productive. 
It factors the hard problem of intelligent behavior into a comparatively tractable one of constrained optimization, at the cost of treating the objective itself as an unanalyzed input.

The reinforcement learning (RL) tradition makes this division explicit. 
In the Markov decision process (MDP) formalism \cite{sutton1998reinforcement}, an agent is a tuple $(\mathcal{S},\mathcal{A},P,r,\gamma)$ in which $\mathcal{S}$ and $\mathcal{A}$ are state and action spaces, $P:\mathcal{S}\times\mathcal{A}\to\Delta(\mathcal{S})$ is the transition kernel, $\gamma\in[0,1)$ is a discount factor, and the reward function $r:\mathcal{S}\times\mathcal{A}\to\mathbb{R}$ is supplied by the designer. 
The agent seeks a policy $\pi:\mathcal{S}\to\Delta(\mathcal{A})$ that maximizes the expected discounted return
\[
J(\pi) \;=\; \mathbb{E}_{\pi}\!\left[\sum_{t=0}^{\infty}\gamma^{t}\, r(s_t,a_t)\right].
\]
Crucially, $r$ is treated as exogenous.
The agent is not asked to evaluate it, only to optimize it. 
An analogous structure holds in supervised learning, where a loss $\mathcal{L}(\theta;\mathcal{D})$ encodes a target distribution chosen in advance, and in game-playing systems, where the win condition is fixed by the game's rules. 

Within these paradigms, the empirical record is striking.
The universal approximation theorem \cite{hornik1989multilayer} established that neural networks can represent any continuous function, removing expressivity as a principled obstacle.
Empirical milestones such as AlexNet \cite{krizhevsky2017imagenet} demonstrated that gradient-trained deep networks could decisively outperform hand-engineered features on tasks such as image classification.
More recently, reinforcement-learned constructions have refuted long-standing conjectures in mathematics \cite{ju2026ai}, showing that AI can produce results outside the search horizon of human intuition.
These include superhuman play in go, chess, and shogi by a single architecture trained from self-play \cite{silver2018general} and augmented with an environment model without prior knowledge of the rules \cite{schrittwieser2020mastering}. 
In structural biology, \cite{jumper2021highly} reduced a long-standing open problem to a supervised regression against a curated objective. 
In language modeling, the next-token cross-entropy loss \cite{radford2019language} and its post-training extensions via reinforcement learning from human feedback \cite{christiano2017deep,ouyang2022training} have produced systems whose behavior is shaped almost entirely by the choice of training signal.
These successes, however, share a common structural feature: the objective is given. 
Whatever creativity the designer exercises in selecting $r$, $\mathcal{L}$, or the preference dataset, the learning algorithm itself has no control over its selection over another.

The given-objective assumption begins to strain both practically and conceptually. 
As tasks become more open-ended, e.g., robotic manipulation in unstructured homes or autonomous scientific discovery, hand-specified rewards become brittle, sparse, or misleading.
The space of behaviors implicitly endorsed by a given $r$ is rarely the space its designer intended, a phenomenon termed reward misspecification and reward hacking \cite{amodei2016concrete,krakovna2020specification}.
We note that these pathologies are symptoms of the given-objective paradigm rather than challenges that an autotelic system must answer on the designer's terms.
Once the agent is responsible for its own objective, the question is no longer whether its behavior matches an external specification but whether the mechanism that produces the objective is itself coherent.

A handful of research programs have, from quite different starting points, asked the system itself to take over part of the design loop. 
Self-referential learners \cite{schmidhuber2007godel,zhang2025darwin,zhang2026hyperagents} allow an agent to rewrite any part of its own code, including its objective, whenever it can prove the rewrite increases expected utility under its current goal. 
The corresponding failure mode is wireheading, in which the agent discovers that the cheapest route to high utility is to rewrite the reward channel itself rather than the world the reward is meant to track \cite{everitt2017reinforcement,majha2019categorizing}.
Wireheading exposes that self-modification under a fixed utility does not deliver goal autonomy, but rather relocates the designer's choice from the reward function to the proof system that licenses self-edits.
Universal AI \cite{hutter2005universal} retains a fixed reward but combines it with a Solomonoff prior over environments. 
This is not a conceptual advancement in agent autonomy but a precise mathematical formulation of optimal reinforcement learning under a universal class of computable environments.
Open-ended evolutionary systems \cite{wang2020enhanced} and quality-diversity methods \cite{lehman2011abandoning,mouret2015illuminating}, dispense with a single objective and instead reward behavioral diversity. 
Each of these programs concerns different edges of the given objective assumption without dispensing with it entirely: a base utility, a universal prior, or a diversity criterion remains stipulated from outside.

The autotelic perspective takes the next step and asks how an agent might come to entertain a particular objective in the first place \cite{colas2022autotelic}. 
The term itself is borrowed from Greek \textit{autós} (``self'') and \textit{télos} (``goal''). 
The word entered psychology through studies of intrinsically motivated behavior and flow \cite{mirvis1991flow}, where it names the pursuit of activities for their own sake, e.g., a child exploring a room, a mathematician investigating a conjecture, an artist sketching without expectation of reward, or a schoolchild asked to write an essay on the `aim in life' and discovering, in the writing, what that aim might be. 
In each case, the agent is not optimizing a target supplied from elsewhere; it is generating, sustaining, and revising its own.
This capacity is not merely a philosophical curiosity. 
Any system that aspires to human-level intelligence, or artificial general intelligence more broadly, must eventually solve the autotelic problem, since humans manifestly do.

In contrast to the standard agent's map $\pi:\mathcal{S}\to\Delta(\mathcal{A})$ conditioned on a fixed objective, an autotelic agent is formalized as a tuple $(\pi,\mathcal{G},\mu)$ in which $\mathcal{G}$ is a space of representable goals and $\mu\in\Delta(\mathcal{G})$ is an endogenous distribution over them. 
The policy is thus goal-conditioned, $\pi:\mathcal{S}\times\mathcal{G}\to\Delta(\mathcal{A})$. 
The designer no longer chooses $r$; the designer chooses the mechanism that produces $\mathcal{G}$ and $\mu$.

Posing the question, however, is not the same as answering it. 
Existing work on goal-conditioned RL, intrinsic motivation, and developmental robotics has produced effective operational mechanisms for generating goals \cite{colas2022autotelic}, but it leaves several deeper questions unaddressed. 
What kind of object is a goal space, and on what grounds may one prior over it be preferred to another? 
Can such a prior be principled rather than merely convenient? 
If the candidate principles, e.g., curiosity, empowerment, and compression, themselves stand in need of justification, then where does the justificatory chain terminate?
And what kind of agent must be presupposed for any of these questions to have an answer?

The remainder of this article is organized as follows. 
We first argue that intrinsic motivation, despite its empirical success, does not by itself constitute autotelesis (Section~\ref{sec:intrinsic}). 
We then show that any attempt to specify a prior over goals smuggles in substantive assumptions, regardless of how neutral it appears (Section~\ref{sec:prior}). 
We consider whether goal spaces can instead be discovered from the agent's causal coupling to its environment.
This, in turn, forces the introduction of a self (Section~\ref{sec:embedded}). 
The remaining sections examine the consequences of that move within the framework of embedded agency (Section~\ref{sec:embedded}) and the ensuing relativization of the self (Section~\ref{sec:dissolution}).
We discuss extensions to quantum, philosophy, and agentic AI (Section~\ref{sec:extensions}) before concluding the article (Section~\ref{sec:conclusion}).

\section{Intrinsic motivation}
\label{sec:intrinsic}

The autotelic question, as posed in Section~\ref{sec:goal-problem}, asks $\mathcal{G}$ and $\mu$ might arise from the agent rather than the designer.
The most developed body of work bearing on this question is that of intrinsic motivation.
These are a family of techniques in which the feedback driving learning is computed by the agent itself from its internal states, rather than read off from an environment signal.
Before examining whether these techniques actually answer the autotelic question, it is worth briefly reviewing them.

Intrinsic motivation is motivated in contrast to extrinsic motivation in typical learning formalism.
A supervised learner trained with a loss $\mathcal{L}(\theta;\mathcal{D})$ is a heteronomous system in which every signal shaping its parameters can be traced back to a curated dataset $\mathcal{D}$ and a designer-chosen target.
A reinforcement learner trained with a hand-engineered reward function is structurally similar.
Children, by contrast, do not appear to require labeled examples or numerical rewards in order to acquire competence over their bodies and surroundings.
Their early exploration is better described as a search for regularities and surprises in their own sensorimotor stream \cite{gopnik2012scientific,kidd2015psychology}.
This is conceptually closer to autotelesis than to supervised learning.
The signals that drive behavior are functions of the agent's own model of the world, and the targets that emerge are not handed down from outside.

Intrinsic motivation in artificial agents formalizes this intuition.
The earliest computational proposals reframed novelty and surprise as quantities the agent could compute about its own predictions \cite{schmidhuber1991possibility}, and a now-standard typology distinguishes knowledge-based, competence-based, and morphological sources of intrinsic reward \cite{oudeyer2007intrinsic,mirolli2013intrinsically}.
The reward is defined as $r_{\text{int}}(s_t,a_t,s_{t+1}) = f(\mathcal{B}_t)$, where $\mathcal{B}_t$ is the agent's belief state at time $t$ (typically a parametrized predictive model $\hat{p}_{\theta_t}$, a posterior over models, or a learned representation), and $f$ is a functional that scores epistemic states.
Crucially, the agent's own learning changes $\mathcal{B}_t$, so $r_{\text{int}}$ is endogenous in the sense that no signal external to the agent is needed to compute it.
This structural feature makes intrinsic motivation look similar to autotelic AI.

Common intrinsic motivation metrics include:
\begin{itemize}[nolistsep,noitemsep]
\item Surprise or prediction error: $r = -\log \hat{p}_{\theta_t}(s_{t+1}\mid s_t,a_t)$, which rewards transitions that the current model fails to anticipate.
\item Novelty: $r = -\log \hat{\rho}(s_{t+1})$ or pseudocount-based variants $r = 1/\sqrt{\hat{N}(s)+\epsilon}$, which reward states the agent has rarely visited \cite{bellemare2016unifying}.
\item Information gain: $r = D_{\mathrm{KL}}\big(p(\theta|\mathcal{D}_{t+1}) \| p(\theta| \mathcal{D}_{t})\big)$, the Bayesian update on model parameters induced by the new observation \cite{houthooft2016vime}.
\item Learning progress: $r = \mathcal{L}_{\theta_{t-k}} - \mathcal{L}_{\theta_{t}}$, which rewards regions of the state space where prediction is improving rather than merely poor \cite{oudeyer2007intrinsic2}.
\item Entropy of the visitation distribution: $r = \mathcal{H}\!\big(\rho_\pi(s)\big)$, encouraging maximally uniform coverage of the state space \cite{hazan2019provably}.
\item Empowerment: the agent-centric channel capacity 
$\mathfrak{E}(s) = \max_{p(a^{1:n})} I\left(A^{1:n};S_{t+n}|S_t = s\right)$, which rewards states from which the agent can reach the largest variety of future states \cite{klyubin2005empowerment}.
\item Free energy: $r = -\mathcal{F}(s_t)$, which under suitable assumptions unifies surprise minimization and information seeking within a single variational objective \cite{friston2017active}.
\end{itemize}

A theoretically distinguished member of this family is the knowledge-seeking agent (KSA), the intrinsic counterpart of universal AI \cite{hutter2005universal}.
Where AIXI selects actions to maximize expected discounted reward under a Solomonoff prior $\xi$ over computable environments, KSA variants replace the extrinsic reward with a functional of the agent's posterior over environments \cite{orseau2014universal,orseau2013universal}.
Two canonical KSAs are the Shannon-KSA, which maximizes the expected entropy reduction over the environment posterior, and the KL-KSA, which maximizes the expected Kullback–Leibler divergence between successive posteriors.

In retrospect, intrinsic motivation replaces one designer-specified scalar, $r$, with another, $r_{\text{int}}$.
While this is consequential due to $r_{\text{int}}$ being computed from $\mathcal{B}_t$ rather than being supplied from outside, the goal space $\mathcal{G}$ of the autotelic formulation remains implicit.
The agent has a single, permanent terminal preference (be surprised, gain control, compress, reduce free energy).
There is no $\mu$ in any nontrivial sense, because there is no nontrivial $\mathcal{G}$ over which it could vary.

The next section takes up this question from an algorithmic information theory perspective and shows that an attempt to specify a prior over $\mathcal{G}$, including those that purport to be maximally neutral, smuggles in substantive commitments through the choice of representation.

\section{Resource-driven goal prior}
\label{sec:prior}

This section explores plausible principled priors for the goal space, $\mathcal{G}$, within the autotelic formalism.

A minimal commitment is to take a uniform $\mu$ such that any goal is a priori as worthy as any other.
% The difficulty is that the choice of $\mathcal{G}$ already encodes substantive content.
A uniform measure on goals expressed as Cartesian targets in a workspace differs from a uniform measure on goals expressed as symbolic predicates, which differs again from one over natural-language descriptions \cite{colas2020language}.
Representation is prior to distribution, and uniformity inherits whatever bias the representation contains.
This is the familiar lesson that there is no view from nowhere in measure theory \cite{jaynes1968prior}.
A standard remedy is to invoke a representation-independent prior drawn from algorithmic information theory \cite{li2008introduction}.
Writing $K(g)$ for the Kolmogorov complexity of a goal description $g$, the Solomonoff-style prior $P_{\mathrm{Sol}}(g) \propto 2^{-K(g)}$ yields an ordering that is invariant, up to an additive constant, under change of universal description language \cite{solomonoff1964formal,hutter2005universal}.
Yet the epistemic justification for this prior rests on Bayesian convergence theorems for hypotheses about a fixed computable data-generating process.
Goals are not hypotheses; they admit no truth value, and there is no analog of the data likelihood to drive posterior concentration.
% The complexity prior, transposed from inference to volition, is a stipulation rather than a theorem.
% A weaker, developmental defence---that simple goals scaffold complex ones \textcolor{red}{[Forestier et al., ``Intrinsically Motivated Goal Exploration Processes'']}---is a pragmatic claim about curricula, not a principled claim about preference.

A more promising route is to ground $\mu$ not in the description length of a goal but in a more holistic cost of pursuing it.
An embodied agent is not a Turing machine with unbounded tape; it has a finite power budget, a finite memory footprint, a finite tolerance for error, and a finite amount of wall-clock time before its situation changes.
These constraints can be unified in a tuple of resources that the agent must consider to enact a goal.
Following the LEAST taxonomy as introduced in \cite{sarkar2022qksa}, a representative list is:

\begin{itemize}[nolistsep,noitemsep]
\item Length $L(g)$: the description length of the program that realizes $g$, recovering Kolmogorov complexity as a special case.
\item Energy $E(g)$: the thermodynamic cost of executing that program, bounded below by Landauer's principle \cite{landauer1961irreversibility,wolpert2019stochastic}. 
\item Approximation $\epsilon(g)$: the admissible deviation between the realized outcome and the nominal goal.
% , formalized by rate-distortion theory \textcolor{red}{[Berger, \emph{Rate Distortion Theory: A Mathematical Basis for Data Compression}]}.
\item Space $S(g)$: the working memory required for representing and pursuing the goal.
% \textcolor{red}{[Sipser, ``A Complexity Theoretic Approach to Randomness'']}.
\item Time $T(g)$: the number of computational steps until the goal is achieved.
\end{itemize}
Each coordinate is a valid prior.
The autotelic question thus has to address ways to combine them.
Several principled aggregations are already known.
For example, Levin complexity \cite{levin1973universal} adds time logarithmically to description length, the speed prior \cite{schmidhuber2002speed} makes fast programs strictly preferred at fixed length, while Bennett's logical depth \cite{bennett1988logical} weights structures by the time required for their shortest program to produce them.
Each prior orders $\mathcal{G}$ differently.
More generally, given a set of resource measures $\{R_i(g)\}_{i\in\mathcal{I}}$ associated with enacting a goal $g$, an autotelic agent can, in principle, define its own resource functional \cite{sarkar2022qksa}
\[
\mathcal{C}_{\alpha}(g) = \alpha\big(R_1(g), R_2(g), \ldots, R_{|\mathcal{I}|}(g)\big),
\qquad
\mu_{\Phi}(g) \propto 2^{-\mathcal{C}_{\alpha}(g)},
\]
where $\alpha:\mathbb{R}^{|\mathcal{I}|}\to\mathbb{R}$ aggregates the individual resource costs into a scalar cost of pursuit.
A trivial instantiation with $\{R_i\} = \{L,E,A,S,T\}$ is a weighted sum over them.
QKSA evolves $\alpha$ itself via genetic programming, rather than only its parameters within a predetermined equational form, to stay non-committal towards any specific prior.
In this view, $\mu$ is not a metaphysical posit but a reflection of the agent's embodiment.
For example, an agent with abundant time and scarce energy prefers a different family of goals than one with the reverse profile.

A complexity ordering, whether by $K$, $\mathrm{Kt}$, or any other resource functional, is operationally useful for curriculum learning.
An agent that has not yet mastered the cheap goals is unlikely to make progress on the expensive ones, so beginning with low-cost goals and progressing to higher-cost ones is a sound scheduling heuristic. 
However, curricular ordering tells the agent in what order to attempt goals given a choice of prior; it does not tell the agent which goals are worth wanting.
The autotelic question essentially outlives resource-driven priors.

The next section explores an alternative route.
Rather than legislating $\mu$, it asks whether $\mathcal{G}$ itself can be discovered from the agent's causal interventions on its environment, and whether the structure so discovered constrains $\mu$ from below.

\section{Causal intervention}
\label{sec:structure}

In this section, we investigate if $\mathcal{G}$ itself can be carved out of the agent's coupling to its environment, and if the structure so discovered constrains $\mu$ from below.
This requires a representation of the environment that treats the agent's interventions as first-class citizens.

The standard probabilistic framing of an MDP, with its transition kernel $P(s_{t+1}| s_t,a_t)$, conflates two structurally different relationships between actions and outcomes.
A purely correlational $P$ records which $(s_t,a_t,s_{t+1})$ tuples co-occur, and a causal one that insists that $a_t$ is an intervention on the dynamics, e.g., via $do$-calculus \cite{neuberg2003causality,spirtes2000causation}.
Correlations, confounded by spurious common causes, falter under distribution shift and counterfactual claims.
An agent whose model of the world is purely correlational cannot, in any principled sense, distinguish what it influences from what it merely co-witnesses.

A causal model can have a constructive payoff for goal generation.
Algorithmic information dynamics \cite{zenil2019algorithmic} formalizes causal influence in algorithmic terms.
A perturbation $\delta$ on a generative program $p$ has causal weight $\Delta K(p,\delta) = K(p\oplus\delta) - K(p)$, where $K$ is the Kolmogorov complexity, and the components of a system whose perturbations move it furthest in $K$-space act as causal levers on its dynamics.

However, the structural lacuna is that controllability specifies what can be done without specifying what should be done.
Formally, it yields $\mathcal{G}$ but leaves $\mu$ undetermined.
Operationally, the agent has learned a causal map of its sphere of influence but has no preference over its destinations.
% Knowing that one can change the temperature does not explain why one should.
Thus, in retrospect, intrinsic motivation subsumes causal-interventional learning.
The agent is implicitly rewarded for acquiring an accurate causal model of what it can affect, an epistemic achievement that leaves the autotelic question intact.

To make progress, the autotelic project requires a principle whose justification appeals not to a higher principle \cite{dewey1939theory,bostrom2012superintelligent,yudkowsky2004coherent} but to the conditions under which there is an agent to do any justifying at all.
A candidate fixed point comes from cybernetics and theoretical biology \cite{ashby2013design,maturana2012autopoiesis,di2005autopoiesis}.
Let's define a viability set $V\subseteq\mathcal{S}$ as the subset of states compatible with the agent's continued existence as the kind of system it is.
A homeostatic agent is one whose primitive objective is the indicator $r_{\text{viab}}(s) = \mathbf{1}[s\in V]$, or equivalently the maintenance of $\mathbb{P}(s_t\in V)\to 1$ as $t\to\infty$.
Under a free-energy formulation \cite{friston2010free}, minimizing expected surprise upper-bounds the divergence from $V$, predictive accuracy lowers the variance of trajectories around $V$, empowerment guarantees a reachable repertoire that can compensate for perturbations, and skill acquisition extends the time horizon over which $V$ can be maintained.
From this primitive, the standard intrinsic motivations reappear as instrumental consequences.
This objective is self-justifying as an agent that fails to satisfy it ceases to exist \cite{carter1974large,dennett1995darwin}.
So the population of agents who can be subjected to the autotelic question is within the population that pursues homeostasis.

Grounding the autotelic question in homeostasis, however, comes at the price of an often-ignored assumption.
The agent must be a delimited part of the world if there is to be a $V$ for it to remain inside.
The next section examines the structural consequences of this commitment.

\section{Embedded agency}
\label{sec:embedded}

As discussed in the previous section, any complete account of autotelic agency must address the boundary between the agent and the environment.
This section examines what such an account looks like, what it adds, and what gaps remain in the autotelic quest.

\subsection*{Markov blankets}

A principled formalism of the internal-external distinction is the Markov blanket \cite{pearl2014probabilistic}.
It partitions the joint state of the universe $x_t$ into internal variables $\mu_t$, boundary variables $b_t$ further decomposed into sensory $s_t$ and active $a_t$ components, and external variables $\eta_t$, such that the conditional independence $p(\mu,\eta|b) = p(\mu| b)p(\eta|b)$ holds at every time step.
The blanket $b$ then screens off internal from external statistics, i.e., any influence of $\eta$ on $\mu$, or of $\mu$ on $\eta$, must be mediated by $b$.
This operational partition allows the internal states $\mu$ to inherit a well-defined dynamics conditional on the blanket, and external states $\eta$ can be modeled only insofar as they leave traces in $b$.
On this construal, the self is not a metaphysical posit but a statistical structure the agent must induce in order to act \cite{kirchhoff2018markov}.
% The emperor's new Markov blankets
% How to knit your own Markov blanket: Resisting the second law with metamorphic minds
% Between pebbles and organisms: Weaving autonomy into the Markov blanket
It refrains from overloading the agent with the freight of cognitive selfhood, like narrative continuity or qualia.

Markov blanket, however, is a probabilistic graphical model with no obvious grounding in physical agents.
Two recent lines of work address this concern.
On one side, one can show that under fairly minimal assumptions on stochastic dynamics on a state space with sparse coupling, statistical Markov blankets coincide with physical interaction surfaces, in the sense that the boundary identified by conditional independence is the same boundary across which energy and information are exchanged \cite{fields2020markov}.
On the other side, the free-energy principle itself can be viewed as a mechanism that induces the partition rather than presupposing it.
Systems that minimize variational free energy under a generative model with sparse couplings tend, asymptotically, to acquire and maintain Markov-blanket structure, so that compartmentalization is an emergent rather than imposed feature of self-organizing dynamics \cite{fields2024free}.
A complementary argument is that any realistic account of an agent-environment coupling must include the observation apparatus inside the system being described, since otherwise the supposed agent-environment cut is paid for by an unanalyzed external observer \cite{fields2016building}.
For our purposes, these results jointly bolster that the partition $(\mu,b,\eta)$ may be taken as a non-arbitrary description of a homeostatic system.

\subsection*{Embedded universality}

The canonical framing in classical decision theory \cite{russell1995modern} treats the agent as an entity that exchanges observations and actions with an environment through a fixed interface defined by the modeler.
The agent's internals, e.g., its memory, compute, and objectives, are taken to be unaffected by anything except via that interface.
The environment is similarly insulated from anything the agent does except via designated actions.

This Cartesian dualism is mathematically convenient but ontologically misleading.
Actual agents are made of the same stuff as their environments, share the same physical laws, and have no privileged port through which information can be metaphysically partitioned from causation.

The Markov-blanket picture already departs from this framing.
The embedded agency program rejects this dualism.
It mathematizes the earlier conceptual framework of second-order cybernetics \cite{von2003self}, which includes the observer in the system being modeled.
In this view, a single dynamical map $x_{t+1} = F(x_t)$ governs the entire universe, and the agent is a subsystem $\mu_t \subset x_t$ whose transitions are produced by the same $F$ that produces everything else.
\cite{demski2019embedded} identifies four difficulties that follow from this formulation, each consequential to the autotelic question.
First, decision theory becomes problematic because the agent cannot evaluate counterfactuals about its own action by imagining itself outside the world; it can only consider alternative continuations of the same $F$.
Second, embedded world models are no longer drawn from a hypothesis space strictly larger than the agent, since the agent is part of what it is trying to model, and naive Bayesian treatments invite paradoxes of self-reference.
Third, robust delegation concerns how a present agent can build successor agents, e.g., future selves, learned subagents, modified policies, whose objectives remain aligned with its own without access to extra-systemic specifications.
Fourth, subsystem alignment concerns whether the optimizers internal to the agent (e.g., search procedures, learned heuristics) share the agent's overall objective, given that they too are subsystems of $F$.
The space-time embedded intelligence framework \cite{orseau2012space} formalizes some portions of this.
It generalizes AIXI \cite{hutter2005universal} by removing the assumption that the agent's computation is gratis and external to the environment, and instead identifies the agent with a substring of the environment trace.
The agent's actions then include modifications of its own physical realization, including modifications that delete it.
Thus, this framework yields that even an agent with universal inductive capabilities and an unboundedly large goal space cannot, in general, optimize a goal whose evaluation requires reading parts of $x_t$ that lie outside its own light cone, since those parts are not, even in principle, accessible to its own computations.

Recent work has sharpened the constraints on embedded agency.
Universal AI assumes an agent that is computationally larger than the environment that contains it, and many of its convergence and optimality theorems break when that assumption is relaxed.
\cite{everitt2017reinforcement,wyeth2025formalizing} catalogues these failure modes.
For example, an embedded universal agent can wirehead by intervening on the very channel through which its reward is computed, can be deceived by sensor tampering, and can fail to converge on the true environment because its hypothesis class fails to contain sufficiently faithful self-models.
\cite{fallenstein2014problems} shows that the most natural way to formalize an agent that endorses successor agents (requiring the successor's objective provably implies the predecessor's) is incompatible with the L\"obian obstruction in any sufficiently expressive theory.
Thus, a fully self-referential autotelic agent cannot, in general, certify that its own future goal-generating mechanism is trustworthy.
In a similar vein, \cite{fallenstein2015reflective} constructs reflective oracles that partially recover a Solomonoff-like hypothesis class for self-modeling agents, but at the price of weakening the guarantees of Solomonoff induction.

Several other works aim to narrow the gap between idealized and embedded formalisms by enriching the model of the environment.
\cite{miles2021markov} extends the MDP framework so that the agent's internal computations are themselves states subject to $P$, accounting for the cost of cognition in the value function rather than an externality.
\cite{lewandowski2026world} examines the situation when the environment is strictly more complex than the agent, showing that the optimal policy in such a regime cannot be identified by enumeration over hypotheses but must instead exploit local regularities.
\cite{elshatlawy2025towards} approaches embeddedness from physics rather than decision theory, treating the agent as a computationally bounded observer of an underlying rule and showing that many features we assume to be properties of the world are in fact properties of the slice an observer of that capacity can resolve.

The collective lesson of these results is that, as soon as the agent is denied a privileged metaphysical position, the fixed goal space $\mathcal{G}$, a stable distribution $\mu$, and a self-model that contains its own justification, i.e., the defining characteristics of an autotelic project, become subject to constraints that have no analog in the Cartesian setting.

In this sense, embeddedness is a necessary ingredient.
Without it, there is no $V$ to defend, no $b$ across which evidence can flow, and no place at which the regress of justifications can terminate.
It is, however, not sufficient.
Embedded agency identifies and constrains the agent, but does not by itself generate goals.
A system with a Markov blanket and a viability set has reasons to act, but no resources for choosing between actions that are equally compatible with viability, and homeostasis is agnostic to the structure of $\mu$ within the viability-preserving region of $\mathcal{G}$.
More importantly, the embedding itself is not unique.
The same dynamics $F$ admits many partitions $(\mu_\alpha, b_\alpha, \eta_\alpha)$ that satisfy the Markov-blanket factorization \cite{krakauer2020information}.
Each such representational choice picks out a different candidate agent, with a different $V$, and therefore a different objective.
The next section discusses this multiplicity.

\section{Dissolution of agency}
\label{sec:dissolution}

In the previous section, we concluded that the same dynamics $F$ admits a family $\{(\mu_\alpha, b_\alpha, \eta_\alpha)\}_\alpha$ of valid Markov-blanket partitions as representational choices.
This section examines this multiplicity to infer what survives of the autotelic project when the self loses its unique claim to ontological priority.
Thereafter, we consolidate the endorsed autotelic framework from these developments.

\subsection*{Coarse-graining criteria}

The partition $(\mu,b,\eta)$ is an adopted coarse-graining that, in turn, defines a particular self.
Each choice $\alpha$ defines a valid agent over the same underlying dynamics, e.g., a cell, an organ, an organism, a colony, or an ecosystem, as a candidate carrier of a Markov blanket.
Formally, these agents can be graded by the mutual information that a coarse-grained slice retains about its own past and future under intervention \cite{krakauer2020information}.
Complementary to this, \cite{levin2019computational} frames the boundary of a self as a quantity that can be enlarged or contracted as the system reorganizes, while integrated information theory \cite{tononi2016integrated} proposes that the privileged partition maximizes intrinsic causal integration $\Phi$.
Coarse-grainings can also be chosen based on how informative the macroscopic causal regularities they reveal \cite{hoel2017map}.
Note that none of these proposals identifies a partition imposed solely by $F$, thus adding additional criteria such as predictive autonomy, computational autopoiesis, or integration, leaving their justification as primitives open.

\subsection*{Non-dual convergence}

Phenomenological traditions argue that the experienced self is a self-model, in which the model's properties can be mistaken for those of what it models \cite{metzinger2004being,buckner2012ego}.
In cognitive science, the self is considered a narrative locus that organizes temporal behavior \cite{dennett2014self,calude2012universe}.
Complementary to these, the enactive program in philosophy of mind treats the self as a sensorimotor pattern that an embodied system enacts rather than possesses \cite{varela2017embodied}.
Thus, the question of personal identity in analytic philosophy terminates in the observation that what we call self is a chain of physical and psychological continuities that admits no further fact beyond those continuities \cite{parfit1987reasons}.

Non-dual contemplative traditions, like Taoism \cite{laozi2000tao} and Madhyamaka \cite{garfield1995fundamental}, hold that the perceived separation between agent and world is a useful but ultimately conventional designation.
Madhyamaka philosophy's doctrine of two truths distinguishes a conventional level at which agents, goals, and boundaries function normally from an ultimate level at which none of these has intrinsic existence.
The Zen koan's non-discursive wisdom captures this as: before enlightenment, one chops wood and carries water; after enlightenment, one chops wood and carries water \cite{suzuki2007manual}.
It portrays the operational point that the realization of dependent boundaries does not abolish the practicality that those boundaries enabled.
The convergence between the formal development in this article and these traditions is not a mystical coincidence.
Both respond to the same structural fact that any boundary placed within a unified dynamics is a representational choice rather than a discovery, and the agent is whatever a chosen boundary delimits.

\subsection*{Instrumentally indispensable fiction}

Let $M$ denote the agent's self-model and let $T$ denote the proposition \emph{the boundary $\mu$ is fundamental}.
Effective action requires $M$ to behave \emph{as if} $T$ holds, since policies $\pi:\mathcal{S}\times\mathcal{G}\to\Delta(\mathcal{A})$ and value functions are defined relative to a delimited self.
Accurate world-modeling, by contrast, requires $M$ to recognize that $T$ is false, since the dynamics admit no such seam.
The autotelic agent is therefore committed to an instrumentally indispensable fiction whose falsity its own inquiry tends to disclose, reminiscent of self-referential paradoxes in formal systems \cite{hofstadter1999godel,hofstadter2007strange}.

The deepest problem with autotelic AI is thus not how the agent generates goals, but how it generates and relativizes the self to which the goals are assigned.
The agent must believe in its own boundary in order to survive, and must see through that boundary in order to understand the world it survives in.
The two demands are simultaneously satisfiable only because they operate at different levels of description.
The first level is constitutive of the agent, the second is constitutive of the agent's model of itself.
The autotelic agent is therefore not a system that has dispensed with the self, but one that has rendered the paradox of an operationally indispensable self and its dispensable ontology.

\subsection*{Consolidated framework}

Consolidating the argument in the article as a single framework, an autotelic agent is the tuple $(\pi,\mathcal{G},\mu,\mathcal{C}_{\alpha},V,b,M)$ in which:
\begin{itemize}[nolistsep,noitemsep]
\item $\pi:\mathcal{S}\times\mathcal{G}\to\Delta(\mathcal{A})$ is a goal-conditioned policy.
\item $\mathcal{G}$ is the goal space, populated not by the designer but by the controllable factors of the agent's causal coupling to its environment.
\item $\mu\in\Delta(\mathcal{G})$ is an endogenous distribution over goals, shaped by a resource functional $\mathcal{C}_{\alpha}$ and constrained by the viability set $V$ that grounds the homeostatic fixed point.
\item $b$ is a Markov blanket inside the global dynamics $F$, identified by but not uniquely picked out by $F$, and $V$ is the set of internal states $\mu$ compatible with sustaining $b$.
\item $M$ is the agent's self-model, which behaves conventionally as if the partition $b$ were fundamental, while ultimately acknowledging that it is one coarse-graining among many.
\end{itemize}

The autotelic question admits no solution in any component in isolation.
Intrinsic motivation supplies $r_{\text{int}}$ but not $\mathcal{G}$ (Section~\ref{sec:intrinsic}).
Resource priors supply $\mu$ but not its justification (Section~\ref{sec:prior}).
Causal interventional learning supplies $\mathcal{G}$ as the controllable factors but leaves $\mu$ undetermined except via homeostasis (Section~\ref{sec:structure}).
Homeostasis supplies a self-justifying fixed point but presupposes an embedded agent (Section~\ref{sec:embedded}).
Embeddedness, finally, individuates the agent at the cost of revealing that the individuation is non-unique (this section).
Each ingredient is necessary, but none is sufficient on its own.
What binds them is the joint operation of $V$, $b$, and $M$, i.e., the agent acts because there is a $V$ to defend, the $V$ is well-defined because there is a $b$ to bound it, and the $b$ is taken as fundamental in $M$ for as long as action is required, while being relativized in $M$ whenever inquiry permits.

\section{Three extensions}
\label{sec:extensions}

The consolidated framework invites three further directions that this section explores.
First, the dynamics $F$ in which the agent is embedded need not be classical.
Replacing $F$ by a quantum dynamical map raises the question of whether the agent-environment boundary is even drawable in classical information-theoretic terms, and gives the relativization of $b$ a physical rather than merely representational status.
Second, the operational/ultimate split in $M$ admits a much closer reading in relation to contemplative philosophy than the scope of the previous section.
We find that recent analyses in \cite{sandved2026there} converge on the conclusion advanced in this article from independent directions.
Third, the framework can be made concrete by instantiating it in a contemporary agentic system, where the policy, the goal generator, and the self-model are all implemented in the same substrate.
We pursue these three extensions in turn.

\subsection{Quantum}

The framework developed in the foregoing sections operates on the assumption that the agent and its environment are classical dynamical systems, with classical transition kernels $P(s_{t+1}|s_t, a_t)$ and classically observable states.
Yet the physical world is fundamentally quantum mechanical.
To replace classical dynamics with a quantum unitary or open-system evolution, we need to revisit the agent-environment boundary, which the foregoing sections treated as a representational choice.

Consider the classical agent-environment partition $(\mu_t, b_t, \eta_t)$ and its Markov blanket $b_t$ through which all influence passes.
In classical dynamics, measurement of $b_t$ is non-invasive, i.e., one can, in principle, read $b_t$ without disturbing the agent or the environment.
The distinction between observation (reading $b_t$) and intervention (modifying $b_t$) is sharp.
In quantum mechanics, this boundary dissolves.
Measurement of $b_t$ via an ancillary measurement apparatus necessarily entangles that apparatus with the system \cite{zurek1991decoherence}, thus, the act of observation constitutes an intervention \cite{brukner2016quantum}.
Besides, the decomposition of a composite system into agent and environment is not representationally unique in the quantum setting.
Distinct Hilbert-space factorizations correspond to different choices of basis and degree of entanglement, and swapping between them is operationally costly \cite{gour2008resource,sarkar2026yaqq}.

These motivate a quantum generalization of the autotelic framework.
The classical MDP transition kernel is replaced by a quantum channel formalized as a completely positive trace-preserving (CPTP) map (or, equivalent representations like Kraus operators, Choi matrices, etc.) \cite{kraus1983states}.
The agent's internal evolution is governed by a Hamiltonian $H_\mu$ that includes its coupling to the boundary $b$, while the environment evolves under $H_\eta$.
The interaction Hamiltonian term couples the two, and the boundary $b$ becomes a quantum channel through which information and coherence flow.
This framework is realized concretely in the quantum MDP formalism.
When the agent performs quantum measurements, the process becomes a quantum observable MDP \cite{barry2014quantum}.
In the open dynamics Lindbladian representation $\dot\rho_t = -i[H,\rho_t] + \sum_k \big(L_k\rho_t L_k^\dagger - \tfrac{1}{2}\{L_k^\dagger L_k,\rho_t\}\big)$, $H = H_\mu + H_\eta + H_{\mu\eta}$ and $\{L_k\}_k$ are jump operators that encode dissipation, monitoring, and control \cite{lindblad1976generators}.
The subset of jump operators serves as the operational coupling across the agent-environment boundary, providing a direct formalization of coupling.
% Snow, Jain, Krishnamurthy, Lyapunov Based Stochastic Stability of a Quantum Decision System for Human-Machine Interaction.
% A Noether theorem for Markov processes; John C Baez, Brendan Fong

In the classical setting, the Markov blanket $b$ is defined by conditional independence: $p(\mu,\eta|b) = p(\mu|b)p(\eta|b)$, however, in the quantum setting, a qubit can be entangled with both $\mu$ and $\eta$ and thus cannot be cleanly assigned to either side.
The partial trace $\text{Tr}_\eta[\rho]$ yields the reduced density matrix of the agent-plus-boundary subsystem, and the conditional independence needs to be stated in terms of the mutual information $I(\mu:\eta|b)$.
Thus, an autotelic agent's self-model $M$ must also account for the degree of entanglement across that boundary, and the trade-off between maintaining coherence and maintaining the boundary itself.
The thermodynamic cost of the agent's computation cannot be pushed outside to an external substrate.
In quantum systems, the agent's computation modifies the environment, thus enforcing embeddedness constraints on policy for goal execution, goal generation, and self-modeling.

This observation extends the dissolution argument of the previous section.
Recall that the classical agent needed to internalize the tension between believing in its own boundary to act effectively, while recognizing that the boundary is a representational choice.
The quantum setting enforces the physicalization of this tension, thereby justifying a quantum information-theoretic ordering of partitions. 
The agent must construct a boundary $b$ that suppresses entanglement between $\mu$ and $\eta$ so that the classical MDP framework is operative for goal generation and self-modeling, yet goal execution produces entanglement and entropy.
By coarse-graining, the agent continuously projects its self-model onto the chosen subspace, in a manner similar to quantum error correction's syndrome measurement and correction, using additional flag gadgets to balance system size, decoding overhead, and correction thresholds.
Furthermore, dissolution in a quantum autotelic agent resurfaces the cosmological implication of a block universe.
We note that the present framework is compatible with global dynamics while remaining agnostic about determinism.
Unitary evolution at the universal level is deterministic, but effective agency at the subsystem level is governed by decoherence, coarse-graining, and Born-rule stochasticity; hence, the relevant notion of indeterminism for autotelic agents is operational rather than metaphysical.

Future work needs to consider formalizing a quantum autotelic agent building on QKSA \cite{sarkar2022qksa} (knowledge-seeking agent for a quantum environment), AIXI-q \cite{catt2020gentle} (quantum acceleration of universal prediction), and QAIXI \cite{perrier2025quantum} (quantum extension of AIXI via Kraus operators).
Milestones towards a theory of quantum general intelligence would need to consider a QKSQA (quantum knowledge-seeking quantum agent capable of maintaining a quantum self-model), EQKSQA (embedded quantum knowledge-seeking quantum agent), and EQKSQAA (embedded quantum knowledge-seeking quantum autotelic agent), with each step compounding layers of quantum constraint and self-reference as explored in this article.

\subsection{Philosophy}

The bipolar status of the agent's self-model mirrors discourses that have long occupied contemplative philosophy, epistemology, and logic.
In this section, we present a sample of these ideas to spark readers' interest in a deeper philosophical exploration.

\cite{sandved2026there} have recently shown that the free-energy principle, when formulated at sufficient generality, arrives at conclusions strikingly similar to those developed here.
Their argument, presented independently and contemporaneously with this work, focuses on how minimizing variational free energy under a generative model induces a self-other boundary that is operationally robust but ontologically unmotivated.
The correspondence between their information-theoretic approach suggests that the operational split is not a peculiarity of the present framework but a general feature of how bounded systems must model themselves within larger dynamics.
Crucially, both approaches reach the same conclusion that the agent remains noncommittal to any permanent boundary, appreciating it as operational rather than fundamental.

The doctrine of shunyata (emptiness) in Madhyamaka Buddhism formalizes precisely this operationality without ontological commitment \cite{nagarjuna1970mulamadhyamakakarika,garfield1995fundamental}.
An autotelic agent echoes the doctrine of two truths, which distinguishes the level at which entities, agents, and boundaries operate pragmatically from the level at which they are found to possess no intrinsic nature.
In the Pyrrhonian skeptical tradition, epoché (suspension of judgment about ultimate reality) coexists with normal practical engagement \cite{vogt2015ancient} leading to ataraxia \cite{massie2018ataraxia}.

The proposed framework, however, explicitly avoids two opposites.
On one end, it rejects the Advaitic thesis captured in the mahāvākyas, e.g., tat tvam asi (that art thou), which is understood as claiming that the agent and environment are ultimately identical in a single undifferentiated consciousness \cite{sivananda1949brahma}.
While superficially similar to the dissolution claim, this view posits a metaphysical ground (Brahman, pure consciousness) that the autotelic framework deliberately refuses.
On the other end, it avoids the Ājīvika doctrine of absolute fatalism via niyati, which denies agency entirely by collapsing the agent into a deterministic flow \cite{dasgupta1932history}.
The neutral monism family of philosophical positions is closer to the framework's commitments.
These assert that neither mind (idealism) nor matter (materialism) is fundamental, but both emerge as perspectives on a more basic substrate.
In Daoism, this appears as the wuji (that which is without polarity), from which all differentiated forms emerge and to which they return, without any fundamental distinction between observer and observed \cite{zi2017dao}.
In Advaitic traditions, nirguna brahman (the attributeless absolute) is understood as the undifferentiated source from which both subject and object crystallize through māyā (apparent differentiation) \cite{tejomayananda2009tattvabodha}.

To answer how an autotelic agent would appear to an external observer, we invoke the concept of līlā (divine play) from Vedantic and Tantric philosophies.
In this view, the universe is the play of consciousness (the unifying dynamics), and agency is not the pursuit of external goals through optimization, but the spontaneous generation of possibilities without ultimate purpose.
Understood in these terms, an autotelic agent does not optimize but plays \cite{reimer2025play}.
It generates goals, explores outcomes, and abandons pursuits not because they have been achieved or deemed unprofitable, but because the play has taken a new form, shaped by the projection of global dynamics onto the emergent autotelics.
Yet from within, its behavior is perfectly coherent and rational.

% The ethical implications of the operational/ultimate split will be explored: if an agent recognizes the boundary as conventional, does it have obligations to other agents that likewise constitute themselves as conventional boundaries? This question bridges the autotelic framework to virtue ethics, consequentialism, and non-dual ethics traditions \textcolor{red}{[Śāntideva, \emph{Bodhisattvacaryāvatāra}]}, \textcolor{red}{[Singer, \emph{Practical Ethics}]]}.

\subsection{Agentic AI}

In this section, we address the practicality of whether this thesis can be instantiated as a concrete engineering program.
We propose one such instantiation using an LLM-based architecture.

Autonomous systems typically inherit the reinforcement-learning template,
\[
\text{State} \rightarrow \text{Action} \rightarrow \text{Reward} \rightarrow \text{Policy Update}
\]
Many LLM agents instantiate this template indirectly through hidden surrogates like scalar scoring, heuristic rankers, preference models, or retrospective approval signals that are effectively used as rewards \cite{colas2021towards,colas2023augmenting,teodorescu2023endless,teodorescu2023codeplay}.
We propose to treat the agent as a stochastic state machine whose primary operation is continuation under constraints.
Formally, at each step $I_{t+1} \sim P_\theta(I|S_t, \mathcal{M}_t)$, where $I_{t+1}$ is a candidate intention, $S_t$ is the agent-environment state, and $\mathcal{M}_t$ is the current self-model (memory, boundary assumptions, and policy context).
The intention is passed through an admissibility filter $\Gamma(I_{t+1}, S_t, \mathcal{C}_{\alpha}, V, \Pi_{\text{safety}}) \in \{0,1\}$.
If $\Gamma=1$, the action is executed and the system transitions to $S_{t+1}$, else the intention is rejected and re-sampled.
The viability set $V$ enters as a hard safety and persistence constraint.
The resource functional $\mathcal{C}_{\alpha}$ enters as a soft or hard budget on token, tool, time, and memory expenditure.
The self-model $M$ is realized as a persistent latent state storing role, commitments, open loops, and boundary assumptions.
The resulting loop,
\[
S_t \rightarrow I_{t+1} \sim P_\theta(\cdot\mid S_t,\mathcal{M}_t) \rightarrow \Gamma \rightarrow S_{t+1}
\]
is closer to online control with constrained sampling than to reward optimization.

Evaluation metrics should separate productivity from the agency's structural properties.
Autotelic behavior can be assessed through: (i) continuation coherence (stability of long-horizon trajectories under perturbation), (ii) boundary maintenance (rate of policy/safety violations), (iii) goal emergence diversity (variety of internally proposed subgoals), and (iv) self-repair capacity (ability to recover from rejected intentions without collapse).

The proposed agentic implementation is the engineering counterpart of the consolidated tuple $(\pi,\mathcal{G},\mu,\mathcal{C}_{\alpha},V,b,M)$, such that, the intention generator parameterizes $\pi$, candidate intentions instantiate local elements of $\mathcal{G}$, admissibility and continuation frequencies induce an empirical $\mu$, runtime budgets instantiate $\mathcal{C}_{\alpha}$, policy and safety constraints instantiate $V$, interface and memory boundaries instantiate $b$, and persistent trajectory state instantiates $M$.
Related empirical precedents appear in process-based control and bounded-rationality, deliberative/reactive hybrids, model-predictive control for language agents, constrained decoding pipelines, and autonomous LLM agent benchmarks \cite{brooks1991intelligence,leike2018scalable,yao2022react,shinn2023reflexion,wang2023voyager,lightman2024let,srivastava2025autotelic}.

\section{Conclusion}
\label{sec:conclusion}

This article attempts to answer how an agent comes to entertain its own goals.
The trail of this inquiry led us from the technical machinery of autotelic learning through homeostasis and embedded agency to the paradoxical status of the self. 
The title, `The Tao of Agency,' deliberately invokes Fritjof Capra's `Tao of Physics' \cite{capra2010tao}, which revealed that the observer cannot be cleanly separated from the observed once quantum mechanics and systems theory are properly understood. 
Traditional AI shares this shortcoming with traditional physics, studying intelligence in isolation by partitioning the world into agent and environment.
Analogously, this article reveals that the duality between agent and environment, and between goal and self, is operationally indispensable but ontologically illusory. 
An autotelic agent must simultaneously believe in its own boundary (to act coherently) and see through that boundary (to understand the world). 
This paradox is not a bug to be fixed but the core insight that transcends the false choice between pure agency (an agent divorced from the world) and pure physics (a world described as if no observer existed).

\bibliographystyle{unsrt}
\bibliography{ref.bib}

\end{document}